\documentclass[final,3p,12pt]{elsarticle}

\usepackage{amssymb}

\usepackage{graphicx}
\usepackage{comment}
\usepackage{amsfonts}
\usepackage{amsmath}
\usepackage{color}
\usepackage{booktabs}
\usepackage{stfloats}

\usepackage{subfigure}
\usepackage[citecolor=green]{hyperref}
\usepackage{nameref}
\usepackage{xcolor}
\usepackage{microtype}

\journal{Pattern Recognition}

\begin{document}

\begin{frontmatter}



\title{Under the Hood of Transformer Networks for Trajectory Forecasting}


\author[label1]{Luca Franco\dag}
\author[label1]{Leonardo Placidi\dag}
\author[label2]{Francesco Giuliari}
\author[label3]{Irtiza Hasan}
\author[label2]{Marco Cristani}
\author[label1]{Fabio Galasso}

\address[label1]{Sapienza University of Rome}
            
\address[label2]{University of Verona}
            
\address[label3]{Inception Institute of Artificial Intelligence}

\begin{abstract}
Transformer Networks have established themselves as the \emph{de-facto} state-of-the-art for trajectory forecasting but there is currently no systematic study on their capability to model the motion patterns of people, without interactions with other individuals nor the social context. This paper proposes the first in-depth study of Transformer Networks (TF) and Bidirectional Transformers (BERT) for the forecasting of the individual motion of people, without bells and whistles. We conduct an exhaustive evaluation of input/output representations, problem formulations and sequence modelling, including a novel analysis of their capability to predict multi-modal futures. Out of comparative evaluation on the ETH+UCY benchmark, both TF and BERT are top performers in predicting individual motions, definitely overcoming RNNs and LSTMs. Furthermore, they remain within a narrow margin \emph{wrt} more complex techniques, which include both social interactions and scene contexts. Source code will be released for all conducted experiments.

\end{abstract}


\end{frontmatter}


\newcommand{\blu}[1]{{\emph{\color{blue} {#1}}}}
\newcommand{\tbf}[1]{{\textbf{#1}}}

\newcommand{\TODO}[1]{\textcolor{red}{TODO: #1}}
\newcommand{\FG}[1]{\textcolor{blue}{FG: #1}}
\newcommand{\MC}[1]{\textcolor{cyan}{MC: #1}}
\newcommand{\IH}[1]{\textcolor{green}{IH: #1}}
\newcommand{\Giuliari}[1]{\textcolor{purple}{Giuliari: #1}}
\definecolor{lucas_orange}{rgb}{1, 0.4, 0}
\newcommand{\Luca}[1]{\textcolor{lucas_orange}{Luca: #1}}
\definecolor{bazaar}{rgb}{0.6, 0.47, 0.48}
\newcommand{\LeoP}[1]{\textcolor{bazaar}{LeoP: #1}}

\section{Introduction}\label{sec:intro}

\

\begin{figure*}[t]
\centering
\includegraphics[width=\textwidth]{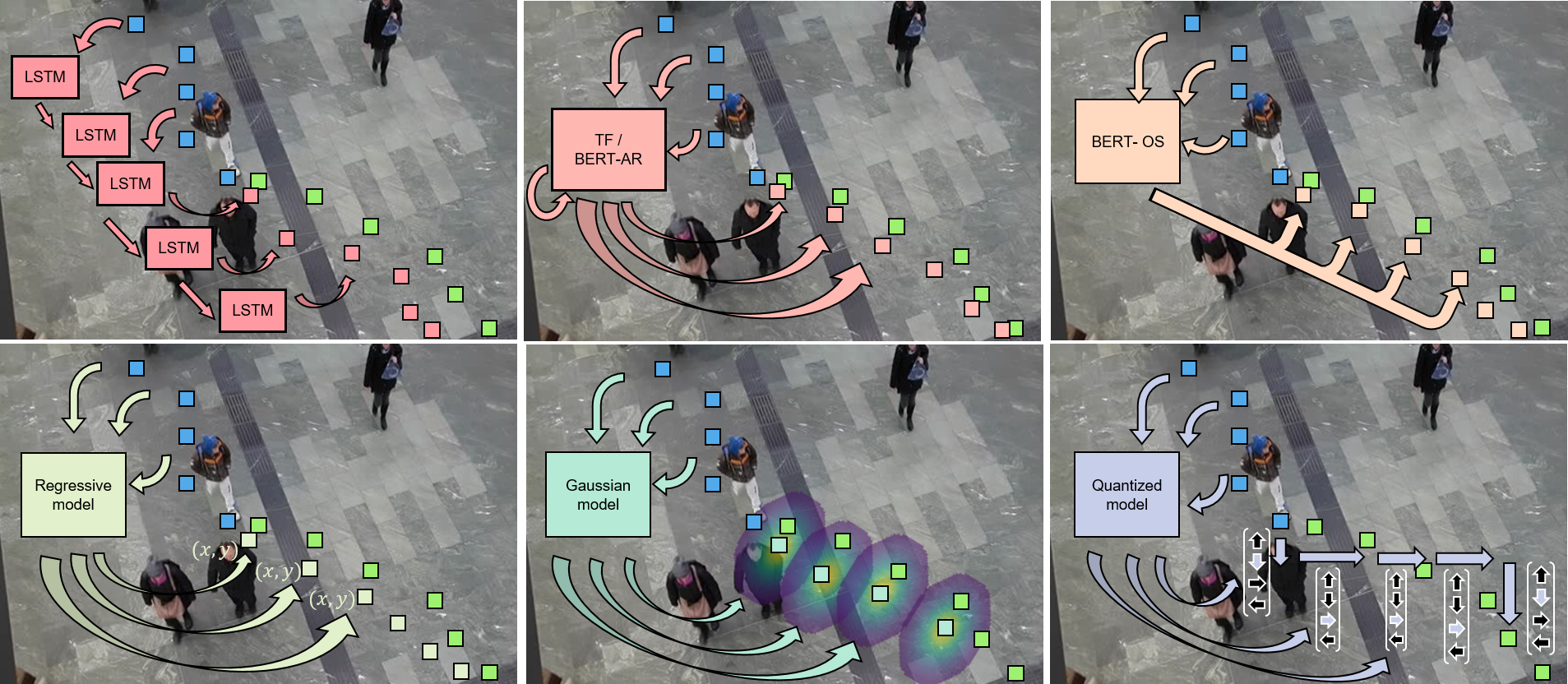}
\caption{Different trajectory forecasting frameworks. \textbf{Top row:} The three model architectures analysed in this paper. \emph{Left:} LSTM recurrently generates future positions by conditioning on previous states. \emph{Middle:} The auto-regressive Transformer predicts future positions one at a time, after observing all past positions concurrently. \emph{Right:} One-shot BERT outputs all future positions at once. \textbf{Bottom row:} The three input/output representations. \emph{Left:} the regressive model directly predicts 2D positions. \emph{Middle:} the Gaussian model outputs the parameters of a Gaussian distribution. \emph{Right:} the quantized model outputs a multinomial distribution over the set of possible movements.}
\label{fig:teaser}
\end{figure*}


The ability to automatically predict the future movements of people is a hot topic in the pattern recognition field, and it is crucial for many modern AI applications, including autonomous driving, human robot interaction and risk management~\cite{lin2021survey}.
Forecasting people is subconscious and effortless for humans, 
but it is not yet solved for machines. In fact, forecasting requires the complex combination of \emph{proxemics} and \emph{scene} reasoning: the former considers that individuals behave differently if they are alone, in a group, or in a crowd.
The latter takes into account the constraints of the scene (e.g.\ people cannot traverse walls).
Additionally, forecasting requires machine learning models to understand both the long- and short-term dependencies in the signals, while being computationally tractable.

Thus far, the way forward has been to use Recurrent Neural Networks (RNN), such as LSTMs and GRUs, to process the sequence of observations and encode them into a single feature that describes the motions. This feature has then been used to predict the future, either by ingestion into an LSTM~\cite{alahi2016cvpr} or into another generative approach~\cite{gupta2018social}. Most work has focused on adding proxemics onto the model, to improve results. The Social-LSTM~\cite{alahi2016cvpr} has started the trend by adding a social pooling module, to take into account the other people in the scene.
Others have focused on integrating the semantics of the scene, to yield future movements which are semantically coherent, considering densely labeled maps~\cite{mangalam2021goals}.

In most recent years, within the deep learning research community, there has been a shift from RNNs to attention-based models such as the Transformer Networks (TF)~\cite{TransformersNIPS17}.
The transition has originated from the Natural Language Processing~(NLP) field, where transformers have yielded an overwhelming increase in performance, esp.\ by the use of bidirectional transformers such as BERT~\cite{BERT19}.
This has impacted trajectory forecasting, too. In fact, TF and BERT were used with social and semantic modules~\cite{yu2020spatio,li2020end,yuan2021agentformer}, reaching top performances on the standard benchmarks. However, so far, there is no direct comparison in literature, between LSTMs and transformers. So it is not clear whether the performance leap is due to the architectural changes, or to improved social and semantic modules.

This work contributes an in-depth study of the transformer models and illustrates how they fare against classical approaches on the standard ETH+UCY benchmark~\cite{pellegrini2009iccv,lerner2007crowds}.
We identify three trends in trajectory forecasting: \emph{single-future}, \emph{multi-future}, and \emph{endpoint-driven} predictions. The first is the classic task of predicting the most likely future motion of a person, given its observed trajectory. The multi-future scenario considers that people may plausibly move into several future directions, i.e.\ they are inherently multi-modal and none of the plausible future motions should be penalized. Endpoint-driven techniques are reminiscent of considering people intentions~\cite{pellegrini2009iccv} and attempt to predict those first~\cite{mangalam2020not,zhao2021you}, afterwards imputing the steps in-between.
%
The ETH+UCY dataset has been chosen to allow for a comparison with the most established approaches, as it is the most commonly used.
We compare transformer models against the corresponding LSTM counterparts, and against the SoA techniques of Social-GAN~\cite{gupta2018social}, Trajectron++~\cite{salzmann2020trajectron++}, and Y-Net~\cite{mangalam2021goals}.

The second main contribution of this work is the analysis of the capabilities of transformers to forecast multiple plausible futures.
We consider for study a modified version of the ETH+UCY dataset, where trajectories have been aligned and clustered together, to emphasize the multi-modal futures of people. Our results confirm that the forecast trajectories of transformers do not collapse into single modes, but they are rather multi-modal, and generalize to diverse endpoint goals, even in the case of test set sequences which differ from the training ones.

This manuscript extends prior work, recently presented in~\cite{giuliari2021trajectory}.
While in that work we had compared performance of Transformer networks and LSTM, in this work we exhaustively explore input-output representations, deterministic and probabilistic problem formulations, we extend the comparison to BERT, we compare auto-regressive and one-shot predictions, and we study the multi-modal predictive capabilities of transformers. \cite{giuliari2021trajectory} has been the first work to have proposed transformer networks for trajectory forecasting. Since then, works adopting transformers have thrived. This manuscript stocks the progress, considering the temporal pattern prediction alone, as the foundation for future further extensions.

The manuscript is organized as follows: Sec.~\ref{sec:method} introduces the considered forecasting modeling choices; Sec.~\ref{Sec:exp} presents experimental evaluation; Sec.~\ref{sec:concl} concludes the work. Next, we discuss related work.

\section{Related work}
\label{sec:prev}




Forecasting people trajectories has been studied for over two decades, and experienced a steady progress from hand-crafted energy-based optimization approaches to data-driven ones. Early approaches have been surveyed in~\cite{morris2008survey}, while deep learning methods (RNNs, LSTMs, CNNs, GANs) are considered  in~\cite{sighencea2021review}; notably, Transformers are totally missing in this panorama, and this paper fills-in this gap. 

We organize the related work in three parts: the first covers Transformers and its bidirectional extension BERT~\cite{BERT19} as general sequence models. The second considers trajectory forecasting approaches which have treated people as independent entities. The third analyzes work modelling the interactions between multiple individuals and between the people and the scene.

\subsection{Transformers and BERT}
Transformer Networks \cite{TransformersNIPS17}, along with their positional encoding schemes, 
have been extensively used  in various fields. Originally focused on NLP applications, Transformers have been successfully deployed in several vision-based tasks, such as image classification, detection and segmentation ~\cite{lin2021survey}. 
Their success has motivated further research, focused on two specific fronts~\cite{lin2021survey}:
\vspace{0.06cm}\\ \noindent \textit{Efficiency.}
Extend the length of the sequences they can process, circumventing the complexity of the self-attention module. Solutions have included sparse attention, linearized attention, memory compression, low-rank attention, attention with prior, improved multi-head attention and decomposing an input sequence into finer segments (see~\cite{lin2021survey} for the specific references).
\vspace{0.06cm}\\ \noindent \textit{Generalization.}
Transformers are agnostic models, and they make few assumptions about the structural bias in the data. 
For this reason, their application to small-scale data is challenging.
Introducing a structural bias and regularizing is beneficial.
Also, recent studies have remarked their capabilities to learn universal language representations by pre-training on large corpora. Notable such cases are BERT~\cite{BERT19} and its evolution, RoBERTa~\cite{RoBERTa19}.

The Transformers' ability to capture long-range dependencies is pivotal for sequence modeling tasks such as trajectory forecasting, and their continuous evolution under different aspects make them attractive and promising.

\subsection{Trajectories as independent entities}
Early work on human path prediction have adopted linear 
 or Gaussian regression models~\cite{quinonero2005unifying}, time-series analysis and auto-regressive models, optimizing for hand-crafted energy functions~\cite{priestley1981spectral}. By contrast, later models have been most successful by the adoption of LSTM~\cite{hochreiter1997long} and RNN models, trained with copious amounts of data.
In particular, LSTM can be employed to regress the future people $x,y$ coordinate positions~\cite{becker2018red,gupta2018social,salzmann2020trajectron++}, or to estimate the Gaussian distributions of the future trajectory coordinates
in order to express the uncertainty associated with the prediction~\cite{alahi2016cvpr,Hasan18}. Unfortunately, Gaussian monomodality could represent a limitation, especially in cases where multiple paths are possible, for example, in crossroads.  

Multimodality has represented a crucial challenge for trajectory forecasting. We divide the literature into two sets: 1) \emph{contextual approaches}: leveraging additional knowledge such as the scene geometry or the presence of other people, i.e., the social context (discussed in the next section); and 2) \emph{agnostic approaches}: multimodal techniques not referring to any additional knowledge.

One of the most known agnostic approaches builds on the determinantal point process (DPP)~\cite{yuan2019diverse}, to sample exhaustively from an arbitrarily multimodal distribution. The main idea is that a variational autoencoder is trained with the influence of a DPP loss so that the samples extracted from the latent space of the trajectories are maximally different. A coarse-to-fine reasoning is adopted in~\cite{bhattacharyya2020haar}, where a  block-autoregressive structure using Haar wavelets is employed to encode long term spatio-temporal correlations. Multimodality can be captured over long time horizons by sampling trajectories at coarse-to-fine spatial and temporal scales. Multi-Choice Learning (MTL) is at the basis of the approach in~\cite{narayanan2021divide}, where a Divide and Conquer (DAC) strategy is used as regularization to MTL, individuating spurious modes where some hypotheses are either untrained in the training process or do not represent any part of the training data. It is worth noting that all of these approaches require a specific module to capture and exploit the multimodality of the data. Here we show that Transformers, without bells and whistles, are capable of capturing the multimodality in their latent space.

\subsection{Social cues and scene context}

The social cues and scene context take forecasting beyond the mere n-dimensional time series prediction, to deal with people.
Social cues make forecasting a choral activity where multiple people, vehicles or, in general, agents are taken into account. This entails collision avoidance and group analysis (people walking together will stay close).
Scene context means including the scene geometry or map to avoid impossible trajectories (e.g. traversing walls) or, in general, to constrain the space of the possible trajectories.

Enabled by the flexibility of the LSTM machinery, best performance has been achieved by modelling the social interaction~\cite{alahi2016cvpr,gupta2018social} among people and the scene context~\cite{salzmann2020trajectron++, zhao2021you}, aided by tracking dynamics~\cite{sadeghian2017tracking} and the spatio-temporal relations among neighboring people~\cite{su2016crowd}.
Some recent works criticised the capability of LSTM to model the human-human interaction~\cite{scholler2020constant,becker2018red,becker2018evaluation}, maintaining that this limits the model generalization capability~\cite{scholler2020constant}. To overcome this issue, they have proposed CNNs to model these interactions~\cite{mangalam2021goals,zhao2021you}.
Y-Net~\cite{mangalam2021goals} takes inspiration from U-Net \cite{ronneberger2015u} and uses CNNs to model both the scene context and the temporal sequence of past motions. STC-Net\cite{li2021spatial} proposes to structure the socio-temporal dependencies in a crowded scene as a graph and to use Graph Neural Networks to predict future motions. 

These social and contextual cues have also been used in conjunction with attention-based models~\cite{sui2021joint,yuan2021agentformer,su2021pedestrian}. In ~\cite{sui2021joint}, images of the scene are fed into the Transformer model to predict jointly the person intention and future trajectory.
AgentFormer~\cite{yuan2021agentformer} models the social interactions of multiple people by aggregating all the temporal sequences by the use of attention.
Since the standard attention disregards the identity of each agent, AgentFormer uses a novel "agent-aware" attention mechanism and attends to the motion of the same person and of others differently.
Despite the great performance, these studies have not thoroughly analysed the contribution of every individual component (social, scene, temporal modelling) in predicting future motion.

In particular, the scene context has been exploited for multimodality (cf.\ the previous section). The hierarchical factorization in~\cite{mangalam2021goals} has partitioned the uncertainty into \emph{epistemic} and \emph{aleatoric}: the former works at a (higher) \emph{endpoint} level, estimating \emph{where} the subject is going to; the aleatoric uncertainty is at a (lower) level, indicating \emph{how} the endpoint is reached, in consideration of all trajectories reaching a given endpoint.
Some earlier work also considers the similar general idea: \emph{Goal-Net}~\cite{cao2020long} works at higher-level, using variational autoencoders to sample an action to be carried out in indoor environments; in~\cite{gu2021densetnt} \emph{target} and \emph{control} uncertainty are synonyms of higher and lower level in an automotive scenario; in the ego-vision automotive setup of~\cite{mangalam2020disentangling}, the higher level is the position on the floor, the lower level the body pose. All these approaches assume that the geometry of the scene is known. 

Recently, several methods have adopted Transformer Networks for sequence modelling \cite{yu2020spatio,li2020end,yuan2021agentformer}.
However, they all have modelled jointly social and temporal aspects.
None of them has exhaustively explored the merits of the temporal-only Transformer which, as it turns out, yields competitive if not better performance than social-temporal models, but with lower complexity.

\section{Forecasting Frameworks}\label{sec:method}

In this section, we model the mere temporal patterns of individual people, without bells and whistles, i.e.\ without consideration of social nor contextual interactions. We discuss input and output representations (Sec.~\ref{sec:inout}), problem formulations (Sec.~\ref{sec:distframework}) and sequence models (Sec.~\ref{sec:seq_mod}).

\subsection{Input and output representations}\label{sec:inout}

Trajectory forecasting stands for observing $T_{obs}$ Cartesian coordinates of the people $\{(x_t,y_t)\}_{t=0}^{T_{obs}-1}$ and predicting the next-future $T_{pred}$ positions $\{(x_t,y_t)\}_{t=T_{obs}}^{T_{pred}+T_{obs}-1}$. At each temporal instant $t$, techniques of trajectory forecasting consider as inputs and outputs either of the representations:
\begin{itemize}
    \item \textit{Positions}: $(x_{t} - x_0 ,\ y_{t} - y_0)$
    \item \textit{Speeds}: $(x_{t} - x_{t-1} ,\ y_{t} - y_{t-1})$
\end{itemize}
So the processed input/outputs at $t$ are either expressed \emph{wrt} a first reference time frame 0, or \emph{wrt} the previous instant $t-1$, which yields speeds.

\textit{Speeds} are \textit{de-facto} the representation of choice of most works, as they provide a more compact coverage of possible input and output values, i.e.\ all values stem from the origin of axes -- zero velocity $(0,0)$, cf.~\cite{becker2018red}. However, \textit{positions} are required when encoding target goals, i.e.\ goals are expressed as endpoints of motion after $T_{pred}+T_{obs}$ instants, starting from $(x_0,y_0)$. Throughout this work, we consider speeds but we also report a comparative evaluation with positions, for the case of Transformer models, in Sec.~\ref{par:TF_models}.

\subsection{Problem formulations}\label{sec:distframework}

In literature, the problem of forecasting trajectories has been formulated in a number of ways. Here we select three main approaches that cover deterministic Vs.\ stochastic, and single-mode Vs.\ multi-modal future predictions.


\subsubsection{Regression}

Regress the output predictions from the inputs, be those positions or speeds (but with consistent input/output pairs).
In this case, future trajectories are deterministic, reported in literature without an indication of uncertainty~\cite{gupta2018social}. And predictions only depict one single plausible future. Training is performed by the using the $L2$ distance between the predicted and the true trajectories as term for the loss.
While seemingly simplistic, the regressive formulation stands surprisingly on par with more complex stochastic and probabilistic approaches (See Secs.~\ref{par:TF_models}, \ref{sec:comp-soa} for the detailed evaluation).

\subsubsection{Unimodal Gaussian distribution}

Trajectory forecasting is formulated as estimating the mean vector $\mu_t$ and the covariance matrix $\Sigma_t$ of a unimodal Gaussian distribution at each time $t$. Inputs are multivariate $(x,y)$ positions or speeds. Outputs are $(\mu_t, \Sigma_t)$. Forecasting proceeds by using the mean $\mu_t$ as the prediction at the next time $t$. Training adopts the negative log-likelihood loss.

This formulation is unimodal, i.e.\ it allows a single future to be predicted, as it only considers a single Gaussian mode; 
the covariance matrix is generally adduced to explain the model uncertainty, i.e.\ the larger the Gaussian bell, the more uncertain the prediction.
There is literature adopting Gaussians~\cite{alahi2016cvpr}, as well as mixing additional inputs such as the view frustum of people~\cite{hasan2019forecasting}. 

\subsubsection{Quantized multinomial distribution}

This formulates trajectory forecasting as a classification problem, i.e.\ the next motion of a person is represented as a vector of possible quantized $K$ moves, and forecasting stands as predicting the correct next move as the highest scoring class from the $K$. Motions may be positions or speeds. Quantization is accomplished by clustering the training set values into $K$ clusters with K-Means. Once the clusters are learnt, any motion is quantized into a one-hot vector. Training adopts the cross-entropy loss.

Each step in the observed and future trajectory is effectively represented by a multinomial distribution over the quantized motions. The forecasting model is fed the observed steps and it outputs a distribution of likely next motions, step-by-step, auto-regressively.
This probabilistic formulation lends itself to predicting multi-modal futures  (cf.\ our study of multi-modality in Sec.~\ref{Sec:multimodal}).
Although it may incur into quantization problems and it requires more parameters, compared to the previous two formulations, this is the best-performing choice for Transformer models.

\subsection{Sequence Models}\label{sec:seq_mod}

We consider Transformer Networks (TF) and Bidirectional Transformers (BERT), which we compare against Long short-term memory (LSTM). BERT is also explained in its one-shot variant, BERT-OS. 

\begin{figure*}[!t]
\centering
  \includegraphics[height=0.34\textwidth]{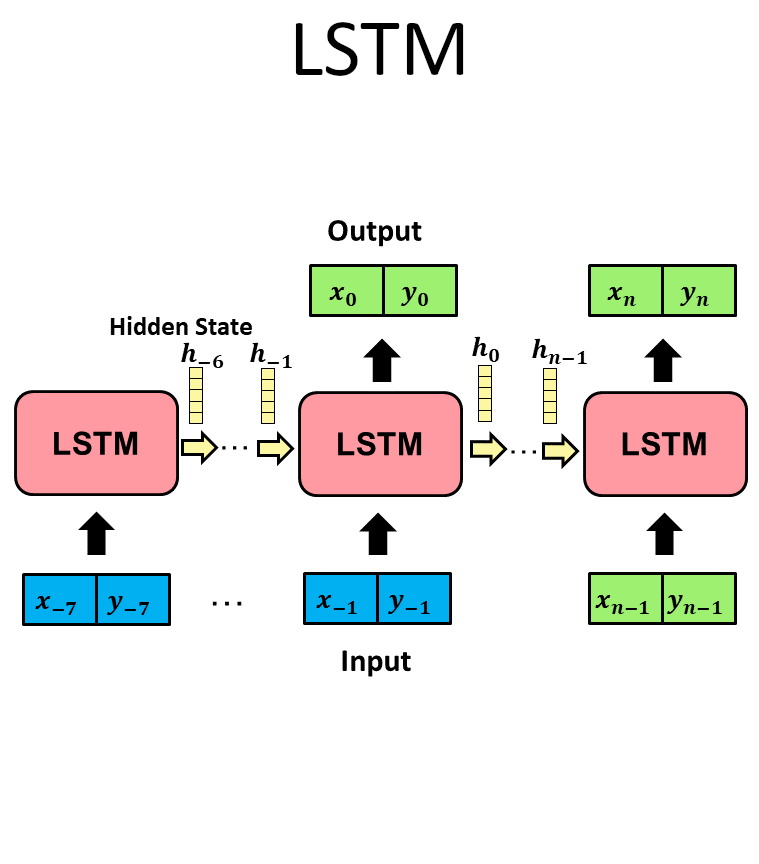}
  \hfill
  \includegraphics[height=0.34 \textwidth]{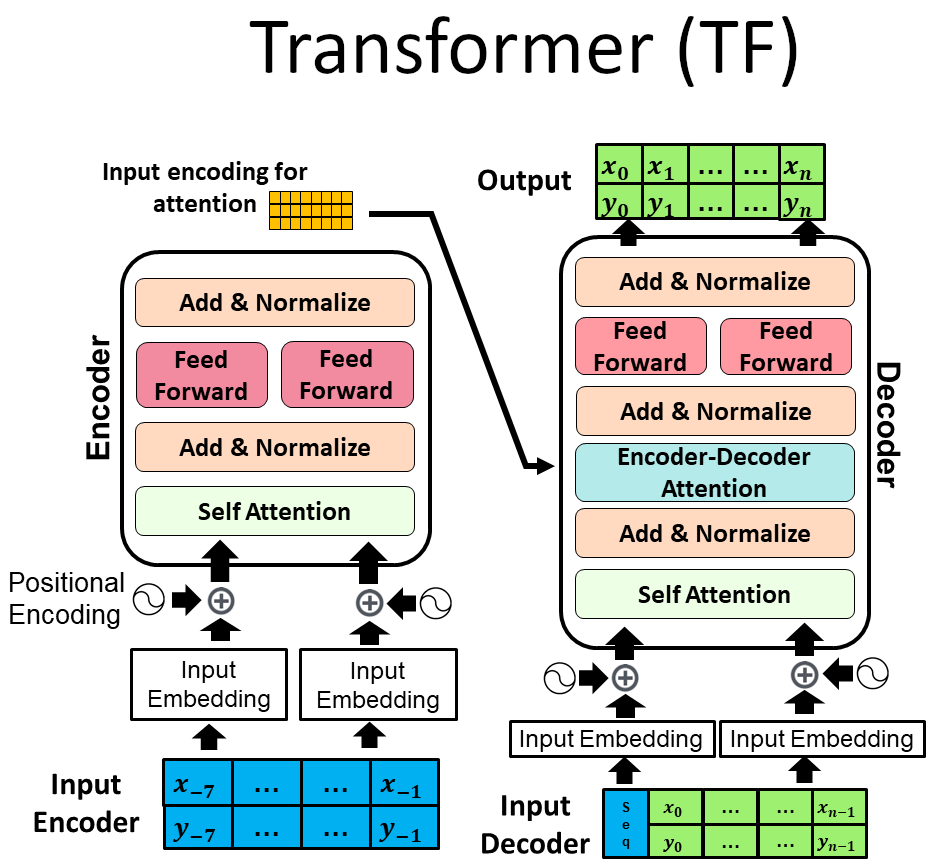}
  \hfill
  \includegraphics[height=0.34 \textwidth]{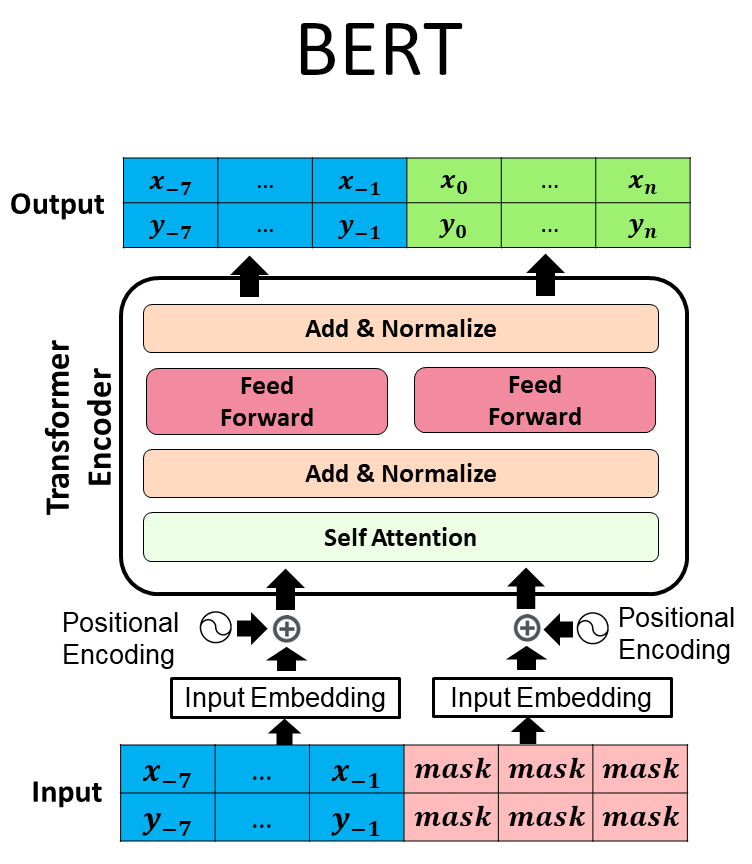}
\caption{Model illustration of LSTM~(\textit{left}), Transformer (\textit{middle}), and BERT (\textit{right}). At each time step, LSTM leverages the current-frame information and its hidden state. By contrast, Transformer utilizes the encoder representation (\textit{yellow}) of the observed input positions (\textit{blue}) and the previously predicted outputs (\textit{green}). BERT uses the Transformer's encoder to directly predict the masked tokens in the input sequence.
}
\label{fig:lstm_tf}
\vspace{-0.2cm}
\end{figure*}

\subsubsection{Transformer Networks (TF)}\label{sec:encdec}

Transformer~\cite{TransformersNIPS17} is the most promising sequence model for trajectory forecasting. It has been introduced in the field by \cite{giuliari2021trajectory} and it now powers the state-of-the-art techniques\cite{yuan2021agentformer,su2021pedestrian,sui2021joint}.
Fig.~\ref{fig:lstm_tf} summarizes the encoder-decoder modular architecture of TF. Each module is repeated \emph{n} times (layers) and consists of 3 basic building blocks: \textbf{i.}\ an attention module, \textbf{ii.}\ a feed-forward module, and \textbf{iii.}\ two residual connections after each of the previous blocks.

The success of TF against LSTM and the former sequence models is mainly adduced to its attention mechanism. Within each attention module, we compute the relevance (via a normalized dot product) of each element of the sequence named "Query" (Q), with all the other sequence entries, named "Keys" (K). Then we re-weight the terms, called now "Values" (V), according to its relevance.
Attention is given by the equation:
\begin{equation}\label{eq:att}
    \textnormal{Attention}(Q,K,V)=\textsf{softmax}\left(\frac{Q K^T}{\sqrt{d_k}}\right) V
\end{equation}

The goal of the encoding stage is to create a representation for the observation sequence, which makes the model \emph{memory}. To this goal, after encoding the $T_{obs}$ inputs, the network outputs two vectors of keys $K_{enc}$ and values $V_{enc}$, to be used by the decoder.
The decoder predicts auto-regressively the future trajectory. At each new prediction step, a decoder query $Q_{dec}$ is compared against the encoder keys $K_{enc}$ and values $V_{enc}$ according to Eq.~\eqref{eq:att} (encoder-decoder attention) and against the previous decoder prediction (self-attention). Cf.\ Sec.~\ref{sec:comp_lstm} for more discussion on the TF memory.

\subsubsection{Bidirectional Transformers (BERT)}\label{sec:bert}

BERT~\cite{BERT19} is an encoder-only transformer network, state-of-the-art in NLP. The supremacy of BERT is due to its larger capacity, being $\sim 2.2\times$ larger than TF, but also to its ability to model bidirectional contextual relations. In fact, BERT introduces the masking mechanism: at training, some words are masked and the model is tasked to predict them from the rest of the sentence, exploiting therefore both the right and the left sides of it. (See also Fig.~\ref{fig:lstm_tf}.)

For trajectory forecasting, we train BERT by providing it the observed parts of trajectories and by masking the futures.
This opens up two approaches to forecasting, which we investigate: \textbf{i.}\ we mask and predict the future next motion only, iterating predictions auto-regressively (as TF does); \textbf{ii.}\ we mask and predict all future instants altogether, forecasting in \emph{one-shot}. We discuss this in more detail in Sec.~\ref{sec:auto_os} and experimentally evaluate in Sec.~\ref{par:TF_models}.

\subsubsection{Comparison with LSTM}\label{sec:comp_lstm}

We note two important differences between TF/BERT and LSTM: the memory mechanism and the positional encoding.
TF maintains the encoded input (cf.\ $K_{enc}$ and $V_{enc}$\,, its \emph{memory} of the observed trajectory) separated from the sequence being decoded (forecast).
Both TF and BERT attend by their self- and encoder-decoder attentions to all input observed portions of the trajectories and to any of the predictions already produced.
TF and BERT have therefore an unlimited memory\footnote{The actual encoded memory is limited to the size of $K_{enc}$ and $V_{enc}$}.
By contrast, LSTM accumulates both the observations and what it has predicted in its hidden state (its gating mechanisms control what is stored or forgotten at any point in time). So the internal state is the limited memory of LSTM.
This contributes to explaining why TF performs better in the long-term horizon predictions, cf. Sec~\ref{Sec:var_length}.

The second significant difference is the positional encoding, cf.\ Fig.~\ref{fig:lstm_tf}. LSTM processes the input sequentially and the order of input positions determines the time flow. It does not therefore require a positional encoding. However, LSTM needs to "unroll" at training time, i.e.\ back-propagate the gradient signal recursively over all input steps. By contrast, positional encoding endows TF and BERT with input and output frames which are time-stamped. This feature, along with the split processing of instants (cf.\ separate feed-forward paths), allows parallelizing the TF and BERT training, which results in more efficient and scalable training.

\subsubsection{Autoregressive VS One-Shot}\label{sec:auto_os}

People trajectories are causal time series. As such, all literature has so far adopted one-directional models, along the temporal dimension. In probabilistic terms, this corresponds to learning the marginal distribution of the next step prediction, conditioned on all past positions, both observed and previously predicted. As a consequence of this modelling, forecasting proceeds \emph{auto-regressively}: predict the next future $t$ given all positions up to time $t-1$, then concatenate $t$ to previous instants to proceed with $t+1$. 

Adopting BERT opens up a novel \emph{one-shot} type of prediction. As introduced in Sec.~\ref{sec:bert}, BERT may be used as a one-directional model, if we only mask the next future instant, both at training and at inference. However BERT natively allows for learning bidirectional contextual relations, when leaving some future instants unmasked. Then, at inference, BERT takes as input all observed positions and it masks all future steps, to predict them altogether. This yields a one-shot prediction of trajectories, for which we dub this BERT-OS. In probabilistic terms, this corresponds to learning the \textit{joint} distribution of each future position, including their mutual dependencies, given all observed positions. This is a novel approach to trajectory forecasting, it yields state-of-the-art performance, and it also allows to effectively leverage intentions, i.e.\ provided endpoint track destinations. (Cf.\ experimental evaluation in Sec.~\ref{par:TF_models} and visual analysis in Sec.~\ref{Sec:multimodal}.)

\subsection{Implementation details}\label{sec:modtrain}

We implement TF according to the original work~\cite{TransformersNIPS17}. In particular, the embedding size is $d_{model}=512$, the number of layers is 6 and the attention heads are 8. The optimizer for the training is Adam, with an initial linear warm-up of 5 epochs.
For BERT, following~\cite{BERT19}, we adopt an embedding size of $d_{model}=768$, 12 layers and 12 attention heads.\\
For the quantized multinomial approach, we set $K=1000$ clusters for K-Means (performance improves slightly with larger $K$ at the cost of larger computation). Prior to this, we densify the manifold by augmenting the training data with uniform random scaling with scales $s\in[0.5,2]$.
Consistently with \cite{graves2013generating}, we normalize the people's speeds by subtracting the mean and dividing by the standard deviation of the train set. 


\section{Experimental Evaluation}\label{Sec:exp}


We experiment on the most-widely adopted datasets for trajectory forecasting, ETH+UCY~\cite{pellegrini2009iccv,lerner2007crowds}. 
We compare sequence models (LSTM, TF, BERT) and forecasting frameworks (regressive, Gaussian, quantized multinomial, see Sec.~\ref{sec:distframework}). We compare against the state-of-the-art for single-track and best-of-20 trajectory predictions. Finally, we analyze in-depth long-term prediction and future multi-modality.

\subsection{The ETH+UCY Benchmark}\label{sec:ethucy}

The ETH+UCY datasets are selected because \emph{all} previous work evaluate on that. These include 5 datasets, collected from 5 different scenes: ETH-univ and ETH-hotel from \cite{pellegrini2009iccv}, UCY-zara01, UCY-zara02 and UCY-univ from \cite{lerner2007crowds}. According to a recent study~\cite{amirian2020opentraj} these scenes span a range of different difficulty, making it an ideal benchmark to test the models in multiple conditions. 

We adopt the evaluation protocol of \cite{alahi2016cvpr}, i.e.\ sample a position every 0.4 seconds and split the entire trajectory into two portions. The first portion, composed of 8 positions (3.2 seconds), is observed. The second portion consists of 12 positions (4.8 seconds) and needs to be forecast.
The evaluation follows a Leave-One-Out (LOO) strategy, i.e.\ models are trained on four datasets and tested on the remaining fifth, then performance is averaged.

Following the literature, we measure performance by Mean Average Displacement (MAD) and Final Average Displacement (FAD). MAD calculates the difference in L2 norm between the 12 future predictions and the respective ground truth in meters. FAD considers only the error at the final point. 

\subsection{Comparing the Transformer models}\label{par:TF_models}

\begin{table*}[t]
\begin{center}
\caption{
Comparison of different combinations of problem formulations (regression, unimodal Gaussian and quantized multinomial distributions) and sequence models (LSTM,
Transformer, BERT-AR). In all cases input/output are speeds, from which positions are computed via integration. MAD and FAD errors are reported in meters.
}
\label{Tab:RegGauss}
\fontsize{8}{8}\selectfont
\resizebox{1\linewidth}{!}{
\begin{tabular}{cccccccccc} \toprule
& \multicolumn{3}{c}{LSTM} & \multicolumn{3}{c}{Transformer} & \multicolumn{3}{c}{BERT - AR} \\
 \cmidrule(lr){2-4}  \cmidrule(lr){5-7}  \cmidrule(lr){8-10} 
\multicolumn{1}{l}{} & Regressive & Gaussian & Qu.(1000) & Regressive & Gaussian & Qu.(1000) & Regressive & Gaussian & Qu.(1000) \\
\midrule
ETH & 1.00/2.00 & 1.21/2.42 & 1.50/2.75 & \textbf{0.93/1.89} & 1.16/2.14 & 1.00/2.12 & 1.05/2.12 & 1.26/2.54 & 1.06/2.24 \\
Hotel & 0.45/0.93 & 0.68/1.43 & 1.53/2.79 & 0.42/0.92 & 0.79/1.41 & 0.30/0.57 & \textbf{0.27/0.53} & 0.40/0.75 & 0.30/0.58 \\
Univ & \textbf{0.53/1.16} & 0.76/1.59 & 1.85/3.39 & 0.55/1.22 & 0.78/1.50  & 0.60/1.23 & 0.68/1.37 & 1.11/2.18 & 0.55/1.19 \\
Zara 1 & 0.42/0.93 & 0.53/1.09 & 1.45/2.64 & \textbf{0.40/0.87} & 0.81/1.40 & 0.44/0.97 & 0.53/0.87 & 0.76/1.56 & 0.44/0.96 \\
Zara 2 &  \textbf{0.33/0.73} & 0.49/1.01 & 1.53/2.80 & 0.34/0.78 & 0.49/0.89 & 0.33/0.74 & 0.43/0.87 & 0.44/0.92 & 0.33/0.73 \\
\midrule
Avg & 0.55/1.15 & 0.73/1.51 & 1.57/2.87 &\textbf{0.53}/1.14 & 0.81/1.48 & \textbf{0.53/1.13} & 0.59/1.15 & 0.79/1.59 & 0.54/1.14 \\
\bottomrule
\end{tabular}
    }
\end{center}
\vspace{-0.3cm}
\end{table*}



In Table~\ref{Tab:RegGauss}, we compare different combinations of problem formulations (regression, unimodal Gaussian and quantized multinomial distributions; cf.\ Sec.~\ref{sec:distframework}) and sequence models (LSTM,
Transformer, BERT-AR; cf.\ Sec.~\ref{sec:seq_mod}).
All models are tasked to process and predict speeds, from which the people trajectories are computed via integration, cf.\ Sec.~\ref{sec:inout}.

\paragraph{Regression}
We first examine the models adopting the regressive formulation, method of choice for much literature, up to recent times~\cite{becker2018red,gupta2018social,salzmann2020trajectron++}.
In this simplest deterministic framework, all techniques achieve comparable performance, i.e.\ LSTM scores 0.55 MAD / 1.15 FAD, TF scores 0.53/1.14, and BERT-AR scores 0.59/1.15. 
The best FAD of TF means that the predicted final position of the person, at the future horizon of 4.8 seconds, is only 1.14 meters, which is typically sufficient for most applications.

\paragraph{Unimodal Gaussian distribution}
All sequence models adopting this problem formulation underperform their counterparts with other formulations. Worst performance is that of TF (0.81/1.48) and BERT-AR (0.79/1.59).
Unimodal Gaussian has been the first go-to choice after the introduction of DNN sequence models~\cite{gupta2018social, alahi2016cvpr, Hasan18}.
The fact that more advanced models perform the worst seems an indication that a Gaussian formulation does not suffice to represent human motion.
Also, the single-Gaussian distribution does not address the inherent multi-modality of people trajectories.

\paragraph{Quantized multinomial distribution}
This formulation allows the more complex models of TF and BERT to shine. The top performer is TF with 0.53/1.13, followed by BERT with 0.54/1.14. By contrast, LSTM severely degrades to 1.57/2.87. Since the models produce a vector of probabilities over each quantum of motion at each time, they are capable of representing multi-modality, too. E.g.\ Approaching an intersection, right and left turns are both feasible futures.
This added skill makes the models more complex, since the goal is now 1000-way classification (see also Sec.~\ref{sec:modtrain}), and it rewards the better sequence models of TF and BERT that can handle it. Notably, only few works~\cite{Chai2019MultiPathMP} adopt the quantized multinomial framework for trajectory prediction in the automotive field, which we instead suggest as the most general and capable for trajectory forecasting.



\subsection{Comparison against State of the Art: Single-future Prediction} \label{sec:soaeval}

\begin{table*}[t]
\begin{center}
\fontsize{8}{8}\selectfont
\caption{
Comparison of \textit{individual} TF and BERT against SoA models on the single-future prediction task. The term "individual" stands for not using \textit{social} nor semantic \textit{map} information. \textbf{Bold} results are best among comparable individual techniques. \textit{Italic} results are best overall, including techniques using social and semantic map information.
}
\label{Tab:ETHUCYdet}
\resizebox{1\linewidth}{!}{
\begin{tabular}{ccccccccccc} \toprule
     &Linear&\multicolumn{5}{c}{LSTM-based}&\multicolumn{2}{c}{TF-based} & \multicolumn{2}{c}{BERT-based}\\
     \cmidrule(lr){2-2}\cmidrule(lr){3-7} \cmidrule(lr){8-9} \cmidrule(lr){10-11}
     &Individual & \multicolumn{2}{c}{Individual} &\multicolumn{2}{c}{Social} & Soc.+ map & \multicolumn{4}{c}{Individual}\\
      \cmidrule(lr){2-2}\cmidrule(lr){3-4} \cmidrule(lr){5-6} \cmidrule(lr){7-7} \cmidrule(lr){8-11}
     &Interpolat.&LSTM&  S-GAN  & Social &  Soc.  & Traj++ & Regress. & Quant & Regress. & Quant \\
     & & \cite{gupta2018social}&  -ind~\cite{gupta2018social}  & LSTM~\cite{gupta2018social}      &   Att.~\cite{kosaraju2019social} & \cite{salzmann2020trajectron++} &  & (1000) & & (1000)\\
     \midrule
  ETH &1.33/2.94       & 1.09/2.94      & 1.13/2.21   & 1.09/2.35 &\emph{0.39}/3.74& \emph{0.39/0.83}& \textbf{0.93/1.89} & 1.00/2.12 & 1.05/2.12 & 1.06/2.24 \\
  Hotel &0.39/0.72       & 0.86/1.91      & 1.01/2.18   & 0.79/1.76&0.29/2.64& \emph{0.12/0.21}& 0.42/0.92  & 0.30/0.57 & \textbf{0.27/0.53} & 0.30/0.58 \\
 UCY &0.82/1.59       & 0.61/1.31      & 0.60/1.28   & 0.67/1.40 &\emph{0.20}/0.52& \emph{0.20/0.44}& \textbf{0.55}/1.22  & 0.60/1.23 & 0.68/1.37  &  \textbf{0.55/1.19} \\
  Zara1 &0.62/1.21       & 0.41/0.88      & 0.42/0.91   & 0.47/1.00 &0.30/2.13& \emph{0.15/0.33}& \textbf{0.40/0.87}   &  0.44/0.97 &  0.53/\textbf{0.87} & 0.44/0.96 \\
  Zara2 &0.77/1.48       & 0.52/1.11      & 0.52/1.11   & 0.56/1.17 &0.33/3.92& 0.14/0.24 & 0.34/0.78  & \textbf{0.33}/0.74 & 0.43/0.87 & \textbf{0.33/0.73} \\
  \midrule
    Avg &0.79/1.59       & 0.70/1.52      & 0.74/1.54   & 0.72/1.54 &0.30/2.59& \emph{0.19/0.41} &\textbf{0.53}/1.14  & \textbf{0.53/1.13} & 0.59/1.15 &  0.54/1.14 \\
\bottomrule
\end{tabular}
}
\end{center}
\vspace{-0.3cm}
\end{table*}

In Table \ref{Tab:ETHUCYdet}, we evaluate the performance of the TF and BERT models, adopting the quantized multinomial formulation, against state-of-the-art techniques, on the task of single-future prediction.

In the table, our proposed TF and BERT are only comparable on a fair ground with techniques which are \emph{individual}, thus functioning without any social nor semantic map cues, namely \emph{Linear}, \emph{LSTM~\cite{gupta2018social}} and the individual version of \emph{Social GAN~\cite{gupta2018social}}. TF and BERT consistently outperform all techniques by a large margin, $\sim$25\% less error on MAD and FAD (TF/BERT Vs. LSTM/S-GAN-ind~\cite{gupta2018social}).

Only techniques leveraging social and semantic maps challenge TF and BERT. The best performer, Trajectron++~\cite{salzmann2020trajectron++}, uses both cues. Still, TF and BERT score a better FAD than some techniques including also the social term, such as Social LSTM~\cite{gupta2018social} and Social Attention~\cite{kosaraju2019social}. This remarkable performance of TF and BERT, \emph{without bells and whistle}, stresses the importance of curating the temporal model well, a strong lever to good performance.

\subsection{Comparison against State of the Art : Multiple-future Predictions}\label{sec:comp-soa}

\begin{table}[t]
\begin{center}
\fontsize{9}{10}\selectfont
\caption{Comparison of the quantized multinomial TF$_{Quant}$ against SoA models, following the best-of-20 protocol. All SoA approaches leverage additional information (social cues and segmented maps), apart from TF$_{Quant}$ and S-GAN-ind~\cite{gupta2018social}. TF$_{Quant}$ outperforms the latter and performs competitively with the current best Y-Net~\cite{mangalam2021goals}.}
\label{Tab:ETHUCY}
\vspace{-0.1cm}
\resizebox{1\linewidth}{!}{
\begin{tabular}{ccc|ccc|cc} \toprule
      & \multicolumn{2}{c}{Social} &  \multicolumn{3}{c}{Social + map} & \multicolumn{2}{c}{Individual}\\
      \cmidrule(lr){2-3} \cmidrule(lr){4-6} \cmidrule(lr){7-8} &   S-GAN      & Trajectron++  & Soc-BIGAT & AgentFormer & Y-Net & S-GAN-ind  & TF$_{Quant}$ \\
    &~\cite{gupta2018social}&~\cite{gupta2018social}&~\cite{salzmann2020trajectron++}&~\cite{kosaraju2019social}& \cite{yuan2021agentformer} & \cite{mangalam2021goals}\\
    \midrule
  ETH        & 0.87/1.62      &  \textbf{0.35/0.77}   & 0.69/1.29& \textbf{0.26}/0.39 & 0.28/\textbf{0.33} & 0.81/1.52 & \textbf{0.61 / 1.12}                    \\
  Hotel                   & 0.67/1.37       & \textbf{0.18/0.38} & 0.49/1.01 & 0.11/\textbf{0.14} & \textbf{0.10/0.14} & 0.72/1.61 & \textbf{0.18 / 0.30}                     \\
UCY                     & 0.76/1.52      & \textbf{0.22/0.48}  & 0.55/1.32 & 0.26/0.46 & \textbf{0.24/0.41}  & 0.60/1.26 & \textbf{0.35 / 0.65}                     \\
Zara1                    & 0.35/0.68       & \textbf{0.14/0.28}& 0.30/0.62 & \textbf{0.15/0.23} & 0.17/0.27 & 0.34/0.69 & \textbf{0.22 / 0.38}                     \\
Zara2                 & 0.42/0.84       & \textbf{0.14/0.30}  & 0.36/0.75 & 0.14/0.24 & \textbf{0.13/0.22} & 0.42/0.84  & \textbf{0.17 / 0.32}                     \\
\midrule
Avg                        & 0.61/1.21   & \textbf{0.21/0.45}   & 0.48/1.00 & \textbf{0.18}/0.29 & \textbf{0.18/0.27} & 0.58/1.18 & \textbf{0.31 / 0.55}     \\               
\bottomrule
\end{tabular}
}
\end{center}
\vspace{-0.7cm}

\end{table}

In Table \ref{Tab:ETHUCY}, we compare the proposed quantized multinomial TF against the state-of-the-art, following the best-of-20 evaluation protocol. In this setting, every method predicts 20 future trajectories and the results are evaluated against the one closer to the ground truth. The TF multiple futures are obtained by sampling from the quantized multinomial distribution.

TF outperforms S-GAN-ind~\cite{gupta2018social}, the sole method in the table using the temporal information just, similarly to TF. This remarks the superiority of Transformers against LSTM models. Notably, TF also outperforms S-GAN~\cite{gupta2018social}, which includes the social term, and Soc-BiGAT~\cite{kosaraju2019social}, which includes both the social and semantic-map terms.
The current best performer is Y-Net~\cite{mangalam2021goals}, using both the social and semantic map terms, as well as their estimated intended goal. This remarks the importance of intentions, as we also note in the analysis of BERT-OS-oracle in the next section.


\subsection{One-shot prediction using BERT}

We turn to BERT-OS and to forecasting all future steps in one-shot, which we use in the analysis of multi-modality in Sec.~\ref{Sec:multimodal}. In Table \ref{Tab:AR_OS}, we compare BERT-OS to BERT-AR first, then analyse input/output representations, and finally discuss the impact of oracle knowledge of the person's intention. 

\paragraph{Auto-regressive Vs.\ one-shot forecasting}
Results in the first column of Table \ref{Tab:AR_OS} (BERT-OS w/ Speeds) are directly comparable with the BERT-AR performances in Table \ref{Tab:RegGauss}. In particular, considering the quantized multinomial formulation, the average performance of BERT-AR (0.54/1.14) degrades slightly for BERT-OS (0.59/1.25). 
We explain this by a two-fold intuition: \textbf{i.}\ next step prediction is the easiest task, but the complete prediction requires processing by all network layers; by contrast, in BERT-OS each future instant prediction is only aided by the outcome of the same layers for other instants; \textbf{ii.}\ proceeding step-by-step aids smooth trajectories and smoothness is in general an effective motion prior.

\begin{table*}[t]
\begin{center}
\caption{
Analysis on the variants for One-shot forecasting using BERT-OS. The first two columns report the results using different input/output representation, namely \emph{speeds} and \emph{positions}. In the last column we report the results of our BERT-OS with the endpoint position provided as part of the input.
}
\label{Tab:AR_OS}

\fontsize{8}{8}\selectfont
\resizebox{1\linewidth}{!}{
\begin{tabular}{ccccccc} \toprule
& \multicolumn{2}{c}{BERT-OS (/w Speeds)} & \multicolumn{2}{c}{BERT-OS (/w Pos)} & \multicolumn{2}{c}{BERT-OS - Oracle (/w Pos)} \\
 \cmidrule(lr){2-3}  \cmidrule(lr){4-5}  \cmidrule(lr){6-7}
\multicolumn{1}{l}{} & Regressive       & Quant (1000)     & Regressive & Quant (1000) & Regressive  & Quant (1000) \\
\midrule
ETH & 1.03/2.04 & 1.02/2.19 & 1.14/2.13 & 1.20/2.27 & 0.54/0.75 & 0.51/0.52\\
Hotel & 0.48/0.88 & 0.35/0.72 & 0.49/1.02 & 0.61/1.11 & 0.16/0.20 & 0.27/0.30\\
Univ & 0.81/1.58 & 0.75/1.52 & 0.75/1.49 & 0.95/1.80 & 0.23/0.17 & 0.40/0.27\\
Zara 1 & 0.51/1.04 & 0.49/1.10 & 0.48/0.97 & 0.63/1.20 & 0.20/0.22 & 0.31/0.31\\
Zara 2 & 0.39/0.79 & 0.33/0.73 & 0.37/0.74 & 0.45/0.88 & 0.17/0.14 & 0.20/0.18\\
\midrule
Avg & 0.64/1.27 & 0.59/1.25 & 0.65/1.27 & 0.77/1.45 & 0.25/0.30 & 0.34/0.32\\
\bottomrule
\end{tabular}
}
\end{center}
\vspace{-0.3cm}
\end{table*}

\paragraph{Speed Vs.\ positions}\label{par:SpeedPos}
Predicting speeds (BERT-OS w/ Speeds) yield better performance than predicting the relative positions of people \emph{wrt} their first frame ((BERT-OS w/ Pos). E.g.\ for the case of quantized multinomial the performance degrades from 0.59/1.25 to 0.77/1.45.
This justifies much literature adopting this type of input/output representation, cf.~\cite{salzmann2020trajectron++,kothari2021human}, and may be explained as follows: \textbf{i.}\ people most commonly move with continuity, e.g.\ by a constant speed, and predicting a constant output is an easier task; \textbf{ii.}\ this implicitly leverages the current bias of most datasets, where continuing to move with the last observed speed provides a strong baseline~\cite{scholler2020constant}.

\paragraph{Oracle intention}
BERT-OS allows to provide as known input any instant of the trajectory. This encompasses the standard problem definition (observe 8 / predict 12 positions), but it also allows to further provide the model the final position (t+12) and forecast (impute) the 11 missing. In Table \ref{Tab:AR_OS}, the oracle performance of BERT-OS is 0.34/0.32, which greatly reduces the error of 0.77/1.45 of the corresponding BERT-OS w/ Pos (it mainly makes sense to provide the final relative position, not the final speed). The importance of the final position motivates addressing prediction of this specific instant by a separate DNN, which is in fact the current state-of-the-art~\cite{mangalam2021goals}, cf.\ Sec.~\ref{sec:comp-soa}. It is surprising that the FAD score of BERT-OS-Oracle is not 0. Since the 12th future position is given, the model might copy it from the input simply. It turns out instead that BERT alters also the inputs 
in order to predict a coherent sequence instead of yielding a bizarre trajectory just for the sake of reaching the final given position, even at the cost of a larger error on the FAD. 
In fact, here BERT behaves as a "denoiser", correcting the given input.


\subsection{Visual Analysis of Transformers Multi-modality}\label{Sec:multimodal}

We investigate the multi-modal capabilities of Transformers.
First, we detail how we processed the ETH+UCY dataset to extract more multi-modal trajectories; then we use BERT-OS for the visual analysis of multi-modality.

\subsubsection{Preparing the multi-modal data}

An ideal multi-modal dataset should include multiple plausible future trajectories for each given observed track. Recently in the literature, some synthetic dataset were proposed~\cite{Liang2020TheGO}, that use simulators to artificially create this paradoxical setting where multiple futures trajectories are observed. When using real world data, this is by definition impossible, as a person will always move following a single path which is the one that will be observed.

In our multi-modal analysis, we focus on the multi-modality already present in ETH+UCY by applying some transformation to enhance it.
We apply rotation and translation to the ETH+UCY datasets such that: \textbf{i.}\ all trajectories originate from the same point, i.e.\ we translate the trajectories so the first observed position of people coincide on the origin of $x,y$-axes; \textbf{ii.}\ all trajectories proceed rightward, i.e.\ we rotate all trajectories so the 8-th observed position lies on the positive \textit{x}-axis.
For this analysis, we fix the UCY-Zara2 as test set since it depicts a sidewalk in front of a storefront where different behaviours can be easily observed e.g., entering the store, window shopping, etc., and consider all other datasets as training set.


The resulting tracks are still coming from real data, but are grouped closer together.
In other terms, by doing so we have densified the dataset by removing two degrees of variability of the trajectories, i.e.\ their origin and their average direction (at least in the observed portion).

\begin{figure}[h]
\centering
\includegraphics[scale=1.5, width=1\linewidth]{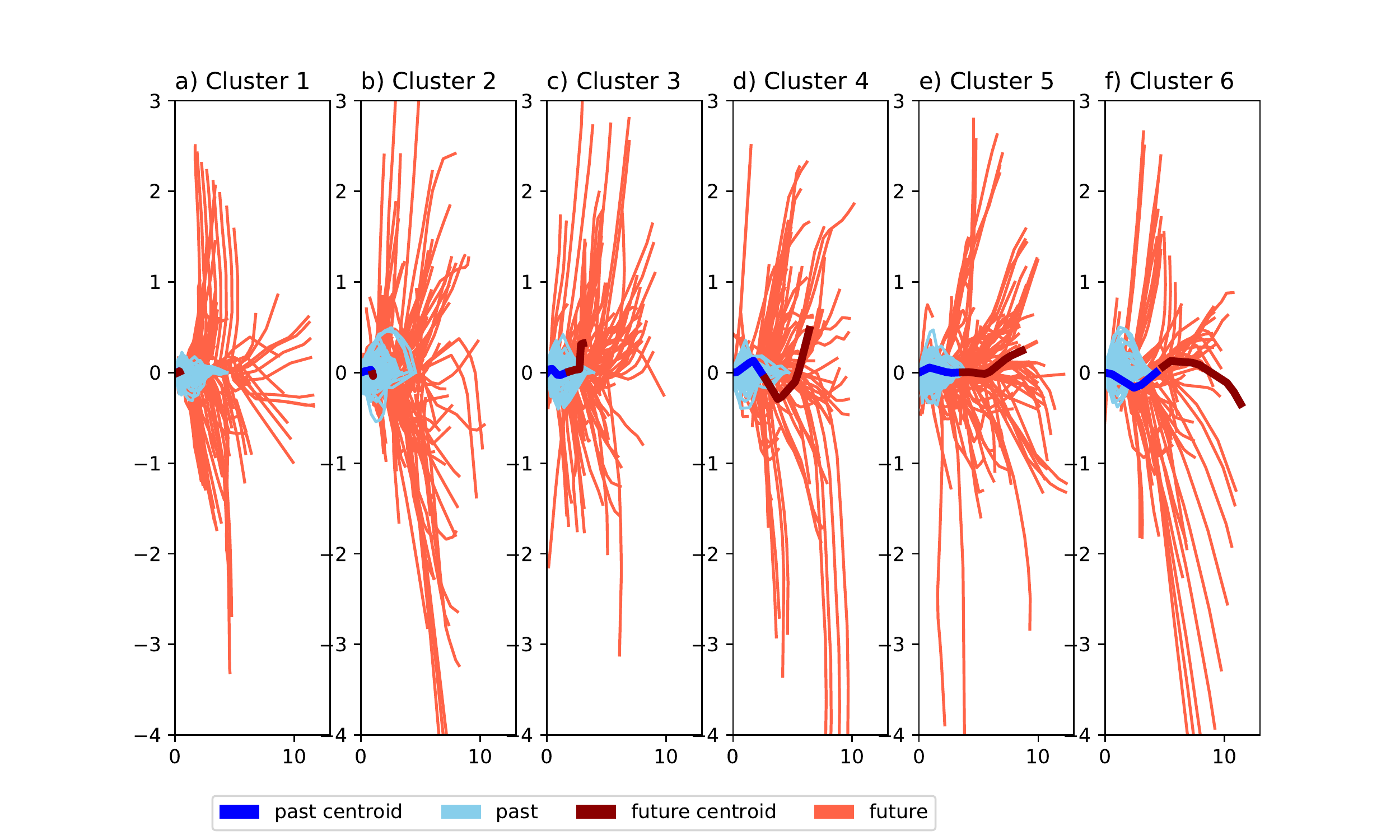}
\caption{Illustration of the 6 main types of people motion in the modified ETH-UCY dataset, clustered according to their speed and variance in motion direction. By construction, all tracks start from (0,0) and cross the \textit{x}-axis at their 8th (last observed) position (see Sec.~\ref{Sec:multimodal} for details and discussion). Tones of blue depict the observed (past) portion of trajectories; tones of red are used for the corresponding futures. The distinct dark blue/red trajectory is the cluster centroid. Clusters are ordered by the average speed, left-to-right and top-to-bottom, and numbered 1-6.
}
\label{fig:distrs}
\end{figure}

In Fig.~\ref{fig:distrs}, we illustrate the 6 main types of people motions in the training set, clustered according to their average speed (\textit{x} coordinate of the 8-th observed position) and the variance in their motion directions. Clusters are numbered 1-6, according to their order in terms of increasing average speed.

Clusters 1-2 (\textit{top-left} and \textit{-mid}) correspond to people who are still or slowly moving during the observed times (\textit{blue} track portions), e.g.\ conversing in groups or looking at window shops. They may either continue to stand still or start moving in various directions in their future (\textit{red} tracks).
Conversely, people of clusters 5-6 (\textit{bottom-mid} and \textit{-right}) walk faster in the observed times. In the future, they may keep the speed or slow down and steer.
In the following study, we focus on cluster 4, which expresses a good variability in terms of diverse final positions of the trajectories.

\subsubsection{BERT-OS with given endpoint}

We challenge Transformer models to predict multiple viable futures from an observed sequence of positions.
The quantized multinomial probabilistic formulation yields best performance and allows multi-modal future predictions.

Two approaches are then applicable to predict all multiple futures: \textbf{i.}\ sampling, and \textbf{ii.}\ imputation.
Research has been devoted to the first~\cite{yuan2019diverse, DLow}. Sampling alone does not suffice as the samples mainly concentrates around the main modes. So efforts have targeted diverse sampling, by introducing diverse sampling modules~\cite{yuan2019diverse} or learnable mappings of latent codes~\cite{DLow}.

We propose to use imputation, since it is based on the mere transformer model.
Our imputation approach exploits the unique capability of BERT-OS: provided with observed positions and a with a tentative goal (i.e.\ final position), the model predicts any missing future position. Adopting imputation is as easy as exploring the predictions of BERT-OS, given a fixed set of observations, while moving the intended person endpoint.


\begin{figure*}
\centering
\includegraphics[width=1\linewidth]{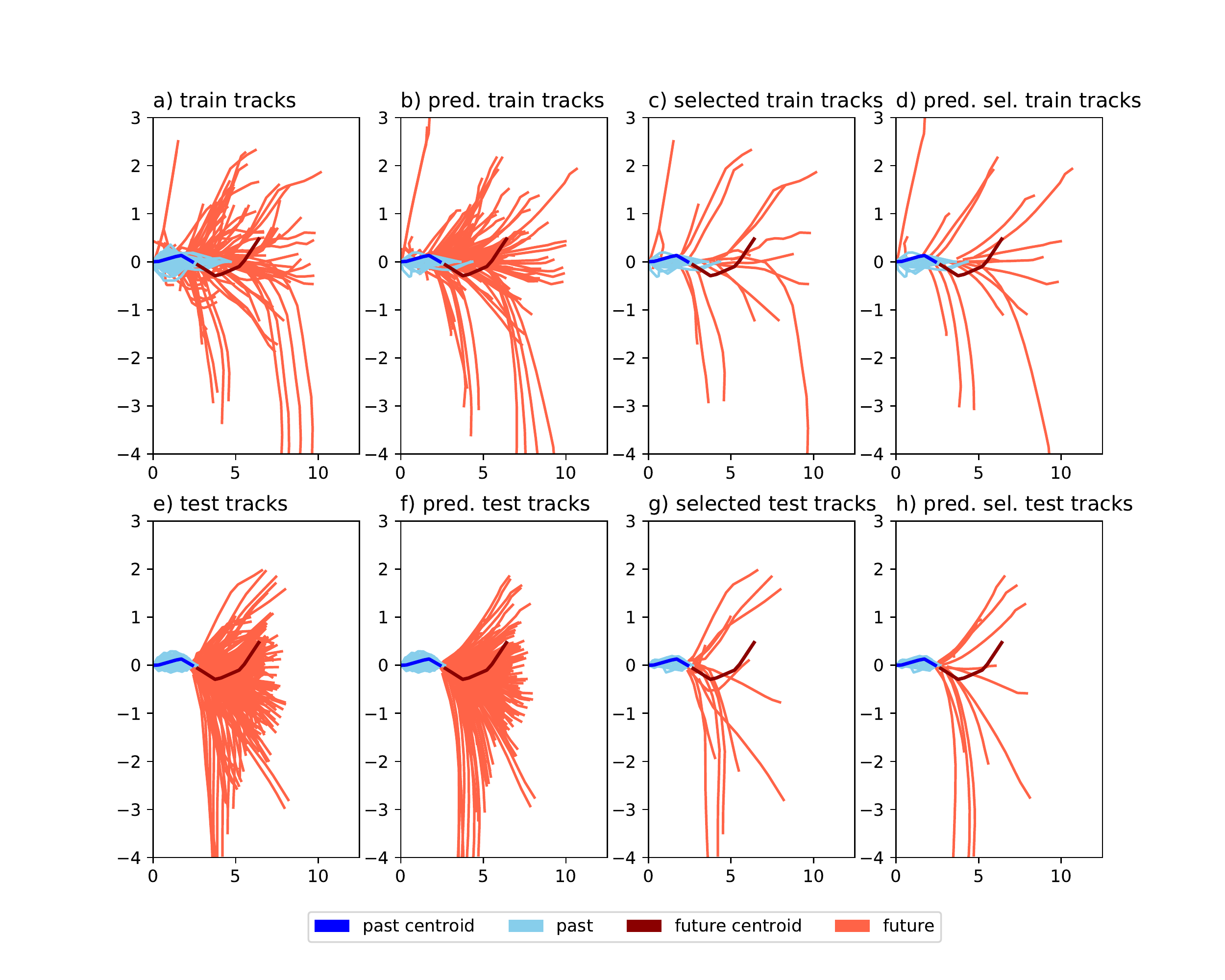}
\caption{
Illustration of the multi-modal prediction capabilities of BERT-OS by imputation on cluster 4 of trajectories from the modified ETH-UCY dataset. The same color convention as in Fig.~\ref{fig:distrs} applies. See Sec.~\ref{sec:vis-analysis} for an in-depth description and discussion.
}
\label{fig:leoc1}
\end{figure*}

\subsubsection{Visual analysis of multi-modality}\label{sec:vis-analysis}

In Fig.~\ref{fig:leoc1}, we showcase the multi-modal prediction capabilities of BERT-OS by imputation for cluster 4. (Similar observations apply to all other motion clusters in Fig.~\ref{fig:distrs}.)

The plots in Fig.~\ref{fig:leoc1}(\textit{a}) illustrate 200 sampled training set sequences from the same cluster. The light blue lines are the 8 observed frames of each sequence and the single darker blue sequence is the cluster centroid; while the red lines are all corresponding parts that need to be predicted.

Fig.~\ref{fig:leoc1}(\textit{b}) illustrates the corresponding predicted training set trajectories.
The depicted red lines are predictions by BERT-OS, given the training set observed positions and the end-points. Discrepancies between (\textit{a}) and (\textit{b}) are due to BERT-OS not managing to learn the complex trajectories of the people, but these are a few, notwithstanding the training regularization of Transformers. For a clearer interpretation, Figs.~\ref{fig:leoc1}(\textit{c}-\textit{d}) reproduce Figs.~\ref{fig:leoc1}(\textit{a}-\textit{b}) with fewer selected tracks. 



Fig.~\ref{fig:leoc1}(\textit{e}) illustrates the different multi-modal distribution of future trajectories in the test set.
The illustrated trajectories are the ones closest (\emph{wrt} the Euclidean distance of the observed track portion) to the centroid from the selected training clusters (i.e., same darker trajectory as in Figs.~\ref{fig:leoc1}(\textit{a}-\textit{d})).

Finally, Fig.~\ref{fig:leoc1}(\textit{f}) illustrates the generalization capability of Transformers to novel multi-modal futures. The figure depicts the predictions of BERT-OS, given the novel test set observed portions of the trajectories and their endpoints (same as in \textit{e}). Note the visual similarity of subplots \textit{e} (GT futures) and \textit{f} (BERT-OS predictions), across the diverse possible endpoints. Figs.~\ref{fig:leoc1}(\textit{g}-\textit{h}) picture fewer trajectories from (\textit{e}-\textit{f}), for a clearer visual impression of the generalizing multi-modal capabilities of Transformers.


Fig.~\ref{fig:leoc1} illustrate that Transformers have the capability to learn successfully, and to mimic well the novel unseen motion patterns across diverse people goals, thus capturing and generalizing multi-modal human motion patterns.

\subsection{Analysis of longer-term forecasting}\label{Sec:var_length}

In this final analysis section, we consider how the forecasting performance changes as the prediction time horizon increases. As in the previous section, we adopt all ETH+UCY datasets as training set, apart from UCY-Zara2, which we retain for testing.

In Table \ref{tab:horizon}, we compare TF and LSTM as the prediction horizon varies from the standard value of 12 frames (4.8s) to 32 frames (12.8s).
Both models are auto-regressive and may extrapolate any number of future steps, but the error is likely to propagate and the longer-term predictions will degrade.

In both models, we observe a gradual degradation of performance. So neither of them will face an abrupt failure.
Notably, TF remains consistently better than LSTM. As argued in  Sec.~\ref{sec:comp_lstm}, this likely results from the more effective memory mechanism of TF: \textbf{i.}\ the observed input sequence, embedded by the encoder, remain accessible at all prediction instants; and the encoded memory is not polluted by any of the predictions. Both aspects are not the case for the LSTM. 








\begin{table}[h] 
\begin{center}
\footnotesize
\caption{
Comparison of LSTM and TF$_\textnormal{Quant}$ when the forecasting horizon is increased from 12 to 32 time steps. See Sec.~\ref{Sec:var_length} for a discussion.\\
}
 \begin{tabular}{ccc}
  \toprule
    {Pred.} & {TF$_\textnormal{Quant}$} & {LSTM}  \\ 
    {} & {MAD / FAD} & {MAD / FAD}\\\midrule
12      & \textbf{0.25/0.59} & 0.50/1.14 \\
16      & \textbf{0.39/0.97} & 0.75/1.74 \\
20      & \textbf{0.57/1.46} & 1.03/2.39 \\
24      & \textbf{0.78/2.10} & 1.33/3.04 \\
28      & \textbf{1.04/2.91} & 1.63/3.69 \\
32      & \textbf{1.35/3.89} & 1.94/4.33\\
\bottomrule
\label{tab:horizon}
\end{tabular}
\end{center}
\end{table}

\section{Conclusions}\label{sec:concl}

In  this  paper  we  have explored exhaustively  the  capabilities  of  Transformer  Networks and of the encoder-only version BERT, in forecasting people motion trajectories. In particular, we have shown the definite superiority of Transformers with respect to LSTMs, as the former overcame the latter across all configurations, at every forecasting time horizon.
When operating on quantized trajectories, the BERT version gave very similar performance to the Transformer Networks. Here, for the first time, we have illustrated the capability of BERT to capture multi-modal future trajectories.
We also have showed that BERT is naturally suited to leverage the estimations of endpoints. All these features should convince to adopt Transformers as workhorses for human trajectory forecasting, and to focus on how to include social proxemics and contextual maps as future research directions.

\section{Acknowledgments}
\vspace{-0.1cm}
This work is partially supported by the Italian MIUR through PRIN 2017 - Project Grant 20172BH297: I-MALL - improving the customer experience in stores by intelligent computer vision, and  by  the  project  of  the  Italian  Ministry  of  Education,  Universities  and  Research  (MIUR)  ”Dipartimenti  di  Eccellenza  2018-2022”.

\bibliographystyle{elsarticle-num} 

\bibliography{compressed_bib}

\begin{thebibliography}{10}
\expandafter\ifx\csname url\endcsname\relax
  \def\url#1{\texttt{#1}}\fi
\expandafter\ifx\csname urlprefix\endcsname\relax\def\urlprefix{URL }\fi
\expandafter\ifx\csname href\endcsname\relax
  \def\href#1#2{#2} \def\path#1{#1}\fi

\bibitem{lin2021survey}
T.~Lin, Y.~Wang, X.~Liu, X.~Qiu, A survey of transformers, arXiv preprint
  arXiv:2106.04554 (2021).

\bibitem{alahi2016cvpr}
A.~Alahi, K.~Goel, V.~Ramanathan, A.~Robicquet, L.~Fei-Fei, S.~Savarese, Social
  {LSTM}: Human trajectory prediction in crowded spaces, in: CVPR, 2016.

\bibitem{gupta2018social}
A.~Gupta, J.~Johnson, L.~Fei-Fei, S.~Savarese, A.~Alahi, Social gan: Socially
  acceptable trajectories with generative adversarial networks, in: CVPR, 2018.

\bibitem{mangalam2021goals}
K.~Mangalam, Y.~An, H.~Girase, J.~Malik, From goals, waypoints \& paths to long
  term human trajectory forecasting, in: ICCV, 2021.

\bibitem{TransformersNIPS17}
A.~Vaswani, N.~Shazeer, N.~Parmar, J.~Uszkoreit, L.~Jones, A.~N. Gomez,
  L.~Kaiser, I.~Polosukhin, Transformer attention is all you need, in: NIPS,
  2017.

\bibitem{BERT19}
J.~Devlin, M.-W. Chang, K.~Lee, K.~Toutanova, Bert pre-training of deep
  bidirectional transformers for language understanding, in: NAACL, 2019.

\bibitem{yu2020spatio}
C.~Yu, X.~Ma, J.~Ren, H.~Zhao, S.~Yi, Spatio-temporal graph transformer
  networks for pedestrian trajectory prediction, in: ECCV, 2020.

\bibitem{li2020end}
L.~Li, B.~Yang, M.~Liang, W.~Zeng, M.~Ren, S.~Segal, R.~Urtasun, End-to-end
  contextual perception and prediction with interaction transformer., in: IROS,
  2020.

\bibitem{yuan2021agentformer}
Y.~Yuan, X.~Weng, Y.~Ou, K.~Kitani, Agentformer: Agent-aware transformers for
  socio-temporal multi-agent forecasting, in: ICCV, 2021.

\bibitem{pellegrini2009iccv}
S.~Pellegrini, A.~Ess, K.~Schindler, L.~Van~Gool, You'll never walk alone:
  Modeling social behavior for multi-target tracking, in: ICCV, 2009.

\bibitem{lerner2007crowds}
A.~Lerner, Y.~Chrysanthou, D.~Lischinski, Crowds by example, in: Computer
  Graphics Forum, 2007.

\bibitem{mangalam2020not}
K.~Mangalam, H.~Girase, S.~Agarwal, K.-H. Lee, E.~Adeli, J.~Malik, A.~Gaidon,
  It is not the journey but the destination: Endpoint conditioned trajectory
  prediction, in: ECCV, 2020.

\bibitem{zhao2021you}
H.~Zhao, R.~P. Wildes, Where are you heading? dynamic trajectory prediction
  with expert goal examples, in: ICCV, 2021.

\bibitem{salzmann2020trajectron++}
T.~Salzmann, B.~Ivanovic, P.~Chakravarty, M.~Pavone, Trajectron++: Multi-agent
  generative trajectory forecasting with heterogeneous data for control, arXiv
  preprint arXiv:2001.03093 (2020).

\bibitem{giuliari2021trajectory}
F.~Giuliari, I.~Hasan, M.~Cristani, F.~Galasso, Transformer networks for
  trajectory forecasting, in: ICPR, 2021.

\bibitem{morris2008survey}
B.~T. Morris, M.~M. Trivedi, A survey of vision-based trajectory learning and
  analysis for surveillance, IEEE Trans. on Circuits and Systems for Video
  Technology (2008).

\bibitem{sighencea2021review}
B.~I. Sighencea, R.~I. Stanciu, C.~D. C{\u{a}}leanu, A review of deep
  learning-based methods for pedestrian trajectory prediction, Sensors (2021).

\bibitem{RoBERTa19}
Y.~Liu, M.~Ott, N.~Goyal, J.~Du, M.~Joshi, D.~Chen, O.~Levy, M.~Lewis,
  L.~Zettlemoyer, V.~Stoyanov, Roberta a robustly optimized bert pretraining
  approach, arXiv:1907.11692 (2019).

\bibitem{quinonero2005unifying}
J.~Qui{\~n}onero-Candela, C.~E. Rasmussen, A unifying view of sparse
  approximate gaussian process regression, Journal of Machine Learning Research
  6~(12) (2005) 1939--1959.

\bibitem{priestley1981spectral}
M.~B. Priestley, Spectral analysis and time series, Academic press, 1981.

\bibitem{hochreiter1997long}
S.~Hochreiter, J.~Schmidhuber, Long short-term memory, Neural computation 9~(8)
  (1997) 1735--1780.

\bibitem{becker2018red}
S.~Becker, R.~Hug, W.~Hubner, M.~Arens, Red: A simple but effective baseline
  predictor for the trajnet benchmark, in: ECCV, 2018.

\bibitem{Hasan18}
I.~Hasan, F.~Setti, T.~Tsesmelis, A.~Del~Bue, F.~Galasso, M.~Cristani, Mx-lstm:
  mixing tracklets and vislets to jointly forecast trajectories and head poses,
  in: CVPR, 2018.

\bibitem{yuan2019diverse}
Y.~Yuan, K.~M. Kitani, Diverse trajectory forecasting with determinantal point
  processes, in: ICLR, 2020.

\bibitem{bhattacharyya2020haar}
A.~Bhattacharyya, C.-N. Straehle, M.~Fritz, B.~Schiele, Haar wavelet based
  block autoregressive flows for trajectories, in: DAGM GCPR, 2020.

\bibitem{narayanan2021divide}
S.~Narayanan, R.~Moslemi, F.~Pittaluga, B.~Liu, M.~Chandraker,
  Divide-and-conquer for lane-aware diverse trajectory prediction, in: CVPR,
  2021.

\bibitem{sadeghian2017tracking}
A.~Sadeghian, A.~Alahi, S.~Savarese, Tracking the untrackable: Learning to
  track multiple cues with long-term dependencies, in: ICCV, 2017.

\bibitem{su2016crowd}
H.~Su, Y.~Dong, J.~Zhu, H.~Ling, B.~Zhang, Crowd scene understanding with
  coherent recurrent neural networks, in: IJCAI, 2016.

\bibitem{scholler2020constant}
C.~Sch{\"o}ller, V.~Aravantinos, F.~Lay, A.~Knoll, What the constant velocity
  model can teach us about pedestrian motion prediction, RA-L (2020).

\bibitem{becker2018evaluation}
S.~Becker, R.~Hug, W.~H{\"u}bner, M.~Arens, An evaluation of trajectory
  prediction approaches and notes on the trajnet benchmark, arXiv (2018).

\bibitem{ronneberger2015u}
O.~Ronneberger, P.~Fischer, T.~Brox, U-net: Convolutional networks for
  biomedical image segmentation, in: MICCAI, 2015.

\bibitem{li2021spatial}
S.~Li, Y.~Zhou, J.~Yi, J.~Gall, Spatial-temporal consistency network for
  low-latency trajectory forecasting, in: ICCV, 2021.

\bibitem{sui2021joint}
Z.~Sui, Y.~Zhou, X.~Zhao, A.~Chen, Y.~Ni, Joint intention and trajectory
  prediction based on transformer, in: IROS, 2021.

\bibitem{su2021pedestrian}
T.~Su, Y.~Meng, Y.~Xu, Pedestrian trajectory prediction via spatial interaction
  transformer network, in: IEEE IV Workshops, 2021.

\bibitem{cao2020long}
Z.~Cao, H.~Gao, K.~Mangalam, Q.-Z. Cai, M.~Vo, J.~Malik, Long-term human motion
  prediction with scene context, in: ECCV, 2020.

\bibitem{gu2021densetnt}
J.~Gu, C.~Sun, H.~Zhao, Densetnt: End-to-end trajectory prediction from dense
  goal sets, in: ICCV, 2021.

\bibitem{mangalam2020disentangling}
K.~Mangalam, E.~Adeli, K.-H. Lee, A.~Gaidon, J.~C. Niebles, Disentangling human
  dynamics for pedestrian locomotion forecasting with noisy supervision, in:
  WACV, 2020.

\bibitem{hasan2019forecasting}
I.~Hasan, F.~Setti, T.~Tsesmelis, V.~Belagiannis, S.~Amin, A.~Del~Bue,
  M.~Cristani, F.~Galasso, Forecasting people trajectories and head poses by
  jointly reasoning on tracklets and vislets, IEEE TPAMI (2019).

\bibitem{graves2013generating}
A.~Graves, Generating sequences with recurrent neural networks, arXiv (2013).

\bibitem{amirian2020opentraj}
J.~Amirian, B.~Zhang, F.~V. Castro, J.~J. Baldelomar, J.-B. Hayet, J.~Pettre,
  Opentraj: Assessing prediction complexity in human trajectories datasets, in:
  ACCV, 2020.

\bibitem{Chai2019MultiPathMP}
Y.~Chai, B.~Sapp, M.~Bansal, D.~Anguelov, Multipath: Multiple probabilistic
  anchor trajectory hypotheses for behavior prediction, in: CoRL, 2019.

\bibitem{kosaraju2019social}
V.~Kosaraju, A.~Sadeghian, R.~Mart{\'\i}n-Mart{\'\i}n, I.~Reid, H.~Rezatofighi,
  S.~Savarese, Social-bigat: Multimodal trajectory forecasting using
  bicycle-gan and graph attention networks, in: NeurIPS, 2019.

\bibitem{kothari2021human}
P.~Kothari, S.~Kreiss, A.~Alahi, Human trajectory forecasting in crowds: A deep
  learning perspective, IEEE Transactions on Intelligent Transportation Systems
  (2021).

\bibitem{Liang2020TheGO}
J.~Liang, L.~Jiang, K.~P. Murphy, T.~Yu, A.~Hauptmann, The garden of forking
  paths: Towards multi-future trajectory prediction, CVPR (2020).

\bibitem{DLow}
Y.~Yuan, K.~Kitani, Dlow: Diversifying latent flows for diverse human motion
  prediction (2020).
\newblock \href {http://arxiv.org/abs/2003.08386} {\path{arXiv:2003.08386}}.

\end{thebibliography}









\end{document}